\documentclass{article}

\usepackage{microtype}
\usepackage{graphicx}
\usepackage{subcaption}
\usepackage{booktabs} 
\usepackage{enumitem}

\usepackage{hyperref}

\usepackage[preprint]{icml2026}

\usepackage{amsmath}
\usepackage{amssymb}
\usepackage{mathtools}
\usepackage{amsthm}

\usepackage[capitalize,noabbrev]{cleveref}

\theoremstyle{plain}

\theoremstyle{definition}

\theoremstyle{remark}

\usepackage[textsize=tiny]{todonotes}

\icmltitlerunning{QANTA 2026 System Description}

\begin{document}

\twocolumn[
  \icmltitle{Task-Specific Multimodal Question Answering Agents via Confidence Calibration and Incremental Reasoning for QANTA 2026}



  \begin{icmlauthorlist}
  \icmlauthor{Nirjhar Das}{cse}
  \icmlauthor{Md. Al-Mamun Provath}{cse}
  \end{icmlauthorlist}

  \icmlaffiliation{cse}{Department of Computer Science and Engineering, Chittagong University of Engineering \& Technology, Chattogram, Bangladesh}

  \icmlcorrespondingauthor{Nirjhar Das}{nirjhardasami@gmail.com}

  \icmlkeywords{Multimodal Question Answering, Quizbowl, Large Language Models, Confidence Calibration, Human-AI Collaboration}

  \vskip 0.3in

]




\printAffiliationsAndNotice{}
\begin{abstract}
We present our submission to the QANTA 2026 shared challenge at the ICML 2026 Workshop on Efficient Multimodal Question Answering (EMM-QA). Quanta evaluates multimodal quizbowl systems that answer pyramid-style questions from incrementally revealed text and accompanying images while operating under realistic efficiency constraints. The challenge consists of two distinct tasks: Tossup questions, which require deciding when to answer under uncertainty, and Bonus questions, which emphasize accurate answer selection and human adoption. To address these differing objectives, we develop a task-specific two-agent architecture. Our Tossup agent utilizes a GPT-4o-mini-class model (referred to as GPT-4.1-mini in the competition logs) with confidence-calibrated answering and a domain-specific numeric reasoning policy that reduces overconfident predictions from isolated quantitative clues. Our Bonus agent uses GPT-4o-class model (referred to as GPT-4.1) with leadin-aware reasoning, structured relational reasoning, and multimodal evidence integration to improve exact answer selection. Rather than relying on a retrieval pipeline or model ensembles, our approach emphasizes efficient reasoning policies and confidence calibration within a hosted-only environment. Our system achieved the highest overall leaderboard score of 0.402, including a Tossup score of 0.238 and a Bonus Effect score of 0.164. The results demonstrate that lightweight, task-specific reasoning strategies can provide strong performance on resource-constrained multimodal question answering benchmarks.
\end{abstract}

\section{Introduction}
The QANTA 2026 shared challenge studies efficient multimodal question answering in the quizbowl setting, where systems answer pyramid-style questions from incrementally revealed text and accompanying images. Multimodal question answering has been extensively studied in visual question answering and multimodal reasoning settings \cite{antol2015vqa}. However, quizbowl-style question answering presents a unique challenge because systems must continuously update their beliefs and decide when sufficient evidence exists to commit to an answer \citep{boydgraber2012besting,wallace2019qanta}. Unlike conventional question answering benchmarks that provide complete context, quizbowl requires systems to reason under uncertainty and determine when sufficient evidence exists to commit to an answer. This setting makes confidence calibration and efficient reasoning as important as answer accuracy.

QANTA consists of two complementary tasks. Tossup questions require systems to decide when to answer as clues are revealed, balancing the benefit of early correct answers against penalties for incorrect predictions. Bonus questions provide complete context and emphasize accurate answer selection, calibration, and usefulness to human teammates. The inclusion of multimodal evidence further requires systems to integrate textual and visual information while remaining computationally efficient.

The distinct objectives of the two tasks motivate a task-specific design. We develop a two-agent architecture consisting of a GPT-4.1-mini-based Tossup agent and a GPT-4.1- based Bonus agent. The Tossup agent focuses on confidence-calibrated answering and robust quantitative reasoning, while the Bonus agent emphasizes leadin-aware reasoning, structured answer selection, and multimodal evidence integration. Our system achieved the highest overall leaderboard score of 0.402, including a Tossup score of 0.238 and a Bonus Effect score of 0.164.

The primary contribution of this work is not a novel model architecture but a collection of task-specific reasoning and calibration strategies that improve performance under the efficiency constraints of QANTA 2026. Specifically, we make the following contributions:
\begin{itemize}[noitemsep,topsep=2pt]
    \item A task-specific two-agent architecture for efficient multimodal question answering
    \item Confidence-calibrated reasoning policies for incremental answering and quantitative reasoning.
    \item An empirical analysis of multimodal reasoning and calibration in the QANTA 2026 shared challenge.
\end{itemize}
Unlike model-centric approaches that improve the underlying language model, our contribution lies in task-specific reasoning and calibration strategies that improve decision quality under competition constraints.

\section{Task Overview}
\label{sec:task}

QANTA 2026 is a shared challenge organized as part of the ICML 2026 Workshop on Efficient Multimodal Question Answering (EMM-QA). The competition evaluates systems on quizbowl-style question answering, a setting that combines incremental reasoning, confidence calibration, and multimodal evidence integration under realistic efficiency constraints. Questions span a wide range of domains, including literature, history, science, fine arts, and current events. Unlike conventional question answering benchmarks that provide complete context before inference, QANTA requires systems to operate on partially revealed information and make decisions under uncertainty.

\subsection{Tossup Task}

Tossups are pyramidal questions revealed incrementally from early, highly distinctive clues to later, more accessible clues. Systems may answer at any point during the clue stream by issuing a buzz. This creates a trade-off between speed and accuracy: answering earlier can increase reward, while incorrect answers incur penalties. Consequently, successful systems must balance answer quality with calibrated decision-making.

Some tossups additionally contain images, requiring systems to integrate visual evidence with incrementally revealed textual clues. Images may include artworks, photographs, maps, diagrams, scientific figures, or OCR-readable text. The central challenge is therefore not only identifying the correct answer, but also determining when sufficient evidence exists to answer confidently.

\subsection{Bonus Task}

Bonuses are multi-part questions that evaluate depth of knowledge and structured reasoning. Each bonus begins with a leadin that defines a shared theme, category, or relationship across constituent parts. Unlike tossups, bonus questions provide complete context before answering and emphasize precise answer selection rather than timing.

Bonus questions may also include images that provide complementary evidence for one or more parts. In the human-AI collaboration setting, AI outputs serve as recommendations to a human captain who makes the final submission decision. As a result, answer accuracy, confidence calibration, and interpretability are all important factors influencing downstream utility.

\subsection{Evaluation Metrics}

Evaluation reflects the decision-making requirements of competitive quizbowl. For tossups, primary metrics include expected score, buzz precision, and average buzz position. These metrics jointly measure answer correctness, calibration, and the ability to answer early without excessive risk.

We evaluate our system using the following metrics as defined by the competition protocol.

\begin{itemize}
    \item \textbf{Win Rate:} The percentage of questions where our system's score exceeds the average score of all other participants.

    \item \textbf{Adoption:} The rate at which the human captain accepted the AI's recommendation.

    \item \textbf{Buzz Position:} The normalized point in the clue stream where the buzz occurred.

    \item \textbf{Bonus Effect:} A composite measure of bonus accuracy and human--AI utility.

    \item \textbf{Calibration:} The alignment between the model's predicted confidence and its empirical accuracy.
\end{itemize}

For bonuses, evaluation emphasizes effect, part-level accuracy, question-level accuracy, calibration, and adoption. In addition, systems are assessed under practical efficiency considerations including inference cost, latency, and token usage, aligning with the workshop's focus on efficient multimodal question answering.

The differing objectives of tossup and bonus questions motivate distinct reasoning and decision-making strategies. This observation guided the design of our task-specific architecture, described in the following section.

\section{System Architecture}
\label{sec:architecture}

The QANTA 2026 competition presents two distinct question-answering tasks with substantially different objectives. Tossup questions require rapid decision-making under uncertainty, where systems must determine not only the correct answer but also the optimal time to answer. In contrast, Bonus questions prioritize answer accuracy, structured reasoning, and usefulness in a human-AI collaboration setting. We therefore adopt a task-specific architecture consisting of two specialized agents rather than a single unified model.

\begin{figure*}[t]
\centering
\includegraphics[width=0.75\textwidth]{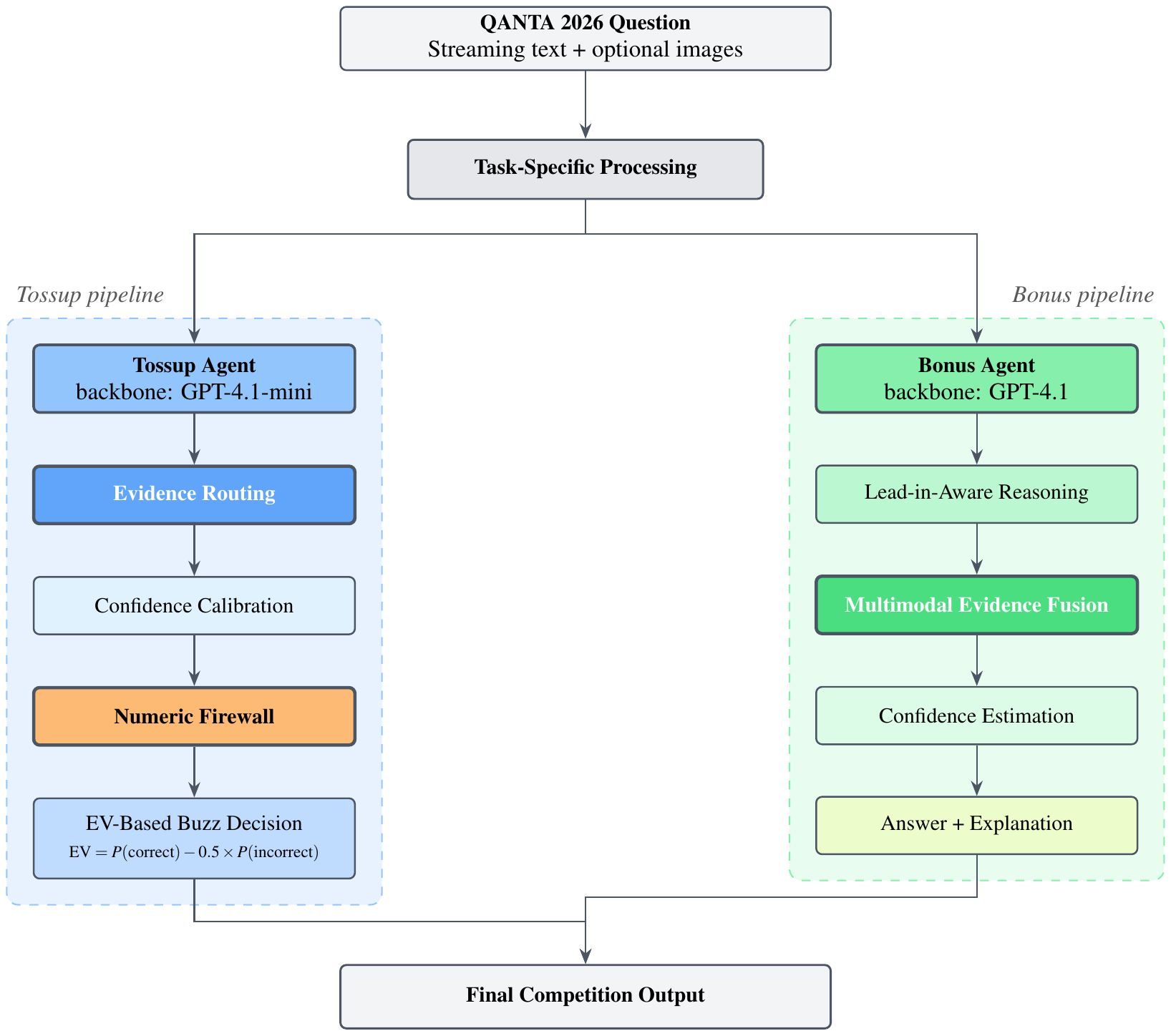}
\caption{
Overview of the proposed QANTA 2026 system. The Tossup agent combines evidence routing, confidence calibration, a Numeric Firewall, and expected-value-based buzzing, while the Bonus agent performs leadin-aware multimodal reasoning and evidence fusion.
}
\label{fig:qanta_architecture}
\end{figure*}

Figure~\ref{fig:qanta_architecture} provides an overview of the proposed system. Incoming multimodal inputs, consisting of textual clues and optional images, are routed according to task type. Tossup questions are processed by a GPT-4.1-mini-based agent optimized for low-latency incremental reasoning and calibrated buzzing decisions. Bonus questions are processed by a GPT-4.1 based agent optimized for precise answer selection, relational reasoning, and human adoption.  Both agents were accessed via the hosted API during the official evaluation window.

\subsection{Design Philosophy}

Our design is guided by three principles: task specialization, confidence calibration, and efficient multimodal reasoning. First, we treat tossups and bonuses as fundamentally different decision-making problems. A single prompting strategy, that performs well on one task may not be optimal for the other. Second, we emphasize calibrated confidence estimates because both competition tracks depend on reliable strategies that leverage visual information when useful while avoiding unnecessary computational overhead.

\subsection{Overall Architecture}

The Tossup agent receives incrementally revealed clues and produces an answer prediction along with an associated verbalized confidence estimate $P(\text{correct})$. This estimate is obtained by prompting the model to provide a numeric certainty score as part of its output stream. These outputs are passed through a confidence-based decision policy that determines whether sufficient evidence exists to issue a buzz. To improve robustness, the system employs a calibration policy that discourages overconfident predictions based solely on isolated numbers, equations, or measurements.

The Bonus agent operates on complete question context and receives the leadin, question text, and any accompanying images. The agent performs leadin-aware reasoning to identify shared themes, categories, or relationships across bonus parts before generating final answers. In addition to answer prediction, the agent produces concise evidence-based explanations intended to facilitate rapid human verification and adoption.

\subsection{Multimodal Reasoning Strategy}

Rather than treating text and images as equally important at all times, our system adopts an evidence-routing strategy. Textual clues are used to generate the initial hypotheses set, while images serve primarily as supporting or disambiguating evidence. Visual information is therefore incorporated selectively when it provides information not already available in the text.

This process is implemented through a late-fusion strategy with cross-checking between modalities. Candidate answers generated from textual evidence are evaluated against available visual evidence before confidence is increased. This approach aims to reduce unnecessary multimodal processing while maintaining the ability to exploit informative images when present.

The resulting architecture aims to provide an effective balance between accuracy, calibration, and computational efficiency, allowing the system to adapt its reasoning strategy to the differing requirements of tossup and bonus question answering.

\section{Tossup Agent}
\label{sec:tossup}

The Tossup task requires systems to answer pyramid-style questions under uncertainty while simultaneously deciding when sufficient evidence exists to commit to an answer. Because incorrect buzzes incur penalties, successful performance depends not only on answer accuracy but also on calibrated confidence estimation. To address these requirements, we employ a GPT-4.1-mini-based Tossup agent optimized for efficient incremental reasoning, confidence calibration, and low-latency inference.

\subsection{Agent Design}

We selected GPT-4.1-mini as the primary Tossup model due to its favorable balance between accuracy, latency, and inference cost. Unlike bonus questions, tossups require repeated evaluation of incrementally revealed clues, making computational efficiency an important consideration. The agent is instructed to behave as an expert quizbowl player, continuously updating its hypotheses as new evidence becomes available while avoiding overcommitment based on generic or weak clues.

The Tossup agent produces two outputs: an answer prediction and an associated confidence estimate. Images are treated as supporting evidence rather than primary information sources. Textual clues drive initial hypotheses generation, while visual evidence is used selectively for disambiguation when it provides unique information not already present in the text.

\subsection{Confidence-Calibrated Buzzing}

A key design decision is the separation of answer generation from buzz decisions. Rather than directly trusting model confidence, the system employs a calibrated confidence gate that determines whether answering is beneficial under the competition scoring rules. Recent work has shown that language model confidence estimates can provide useful signals for uncertainty-aware decision making \cite{kadavath2022language}.

The buzzing policy is governed by an Expected Value (EV) maximization strategy, defined as $EV = P(\text{correct}) - 0.5 \times P(\text{incorrect})$. This formula reflects the competition scoring where correct answers are rewarded and incorrect buzzes incur a penalty weighted at 0.5 of the reward. The buzzing decision can be viewed as a form of selective prediction, where the system chooses whether sufficient confidence exists to commit to an answer \cite{geifman2017selective}.

To ensure a robust margin of safety, we apply a decision gate where a buzz is issued only if the model's verbalized confidence $P(\text{correct}) \geq 0.90$. This conservative threshold was selected based on development results Section~\ref{sec:confidence_calibration} where values below 0.90 frequently corresponded to unstable hypotheses.

\subsection{Numeric Reasoning Guardrails}

A recurring failure mode involved quantitative and scientific questions. The model frequently produced overconfident predictions when exposed to isolated numbers, equations, or measurements without sufficient contextual evidence. These errors often resulted in premature buzzes and unnecessary penalties.

To mitigate this behavior, we introduced a Numeric Firewall, a domain-specific calibration policy designed for mathematics and science questions. The Numeric Firewall is a prompt-level reasoning guardrail that enforces a conditional confidence update. The mechanism prevents the model from increasing $P(\text{correct})$ above a 'safe' baseline when the evidence is limited to isolated numbers or measurements. Confidence is only permitted to exceed the baseline if the numerical data is corroborated by at least one of three specific 'contextual anchors': (1) a named entity, (2) an explicit formulaic context, or (3) multiple independent and consistent clues.  Numeric answers remain valid when explicitly requested by the question or when several independent clues converge on the same conclusion.

This guardrail significantly reduced science-related negations and improved overall calibration, particularly in quantitative domains where numerical patterns alone are often insufficient for reliable identification.

\subsection{Development and Error Analysis}

The final Tossup system emerged through three major iterations. First, the optimization objective was shifted from raw answer accuracy to expected-value maximization, improving the trade-off between reward and risk. Second, the Numeric Firewall was introduced to address quantitative reasoning failures. Third, confidence calibration was refined through position-aware thresholding and adoption of a more  conservative decision threshold.

Common failure mode included over-buzzing on generic clue fragments, confusion between closely related entities, and excessive confidence on isolated numerical evidence. These issues were mitigated through calibrated confidence gating, stronger emphasis on distinctive clues, and the introduction of domain-specific reasoning guardrails.

The resulting system achieved a Tossup score of 0.238 and contributed substantially to the highest overall leaderboard score obtained during the competition.

\section{Bonus Agent}
\label{sec:bonus}

Unlike tossups, bonus questions prioritize answer accuracy, structured reasoning, and usefulness in a human-AI collaboration setting. Bonus questions provide complete context before answering and often requires reasoning over shared themes, categories, or relationships introduced through a leadin. To address these requirements, we employ a dedicated GPT-4.1 based Bonus agent optimized for leadin-aware reasoning, multimodal evidence integration, and human adoption.

\subsection{Agent Design}

We selected the GPT-4.1 model for bonus question answering due to its stronger reasoning, calibration, and disambiguation capabilities. While GPT-4.1-mini provided a favorable efficiency profile for tossups, bonus questions place greater emphasis on answer quality and structured reasoning than on low-latency decision making. In practice, the full model exhibited more reliable behavior on ambiguous questions, relational reasoning tasks, and multimodal examples.

The Bonus agent is prompted as an answer retrieval and disambiguation system rather than an open-ended reasoning assistant. Outputs are intentionally short, structured, and confidence-calibrated. For each question, the model produces an answer, a confidence estimate, and a concise evidence-based explanation intended to support rapid human verification.

\subsection{Leadin-Aware Reasoning}

A central component of the Bonus agent is leadin-aware reasoning. Bonus questions frequently contain a leadin that establishes a shared theme, category, or relationship across multiple parts. Ignoring this context can lead to inconsistent answer selection or failures in relation reasoning.

The Bonus agent therefore processes the leadin before evaluating individual parts. The leadin is used to constrain the candidate answer space and guide subsequent reasoning. This design encourages consistency across bonus parts while improving answer precision in cases where multiple plausible entities remain under consideration.

\subsection{Multimodal Evidence Integration}

The Bonus agent adopts an evidence-routing strategy for multimodal reasoning. Textual clues are used to generate the initial hypotheses set, while visual information is incorporated only when it contributes unique evidence. Images are therefore treated as evidence rather than decoration and may provide OCR text, diagrams, maps, scientific structure, artistic styles, or other identifying signals.

The multimodal reasoning pipeline follows a sequential late-fusion process: (1) Hypotheses Generation: The agent uses textual clues to generate a set of candidate answers. (2) Visual Cross-Checking: The agent evaluates the image (seeking OCR text, diagrams, or artistic styles) against the text-based candidates. (3) Confidence Adjustment: If the image supports or refines a candidate, confidence is increased; if the image contradicts the text-derived hypotheses, the candidate is discarded or confidence is lowered.

This approach reduces premature multimodal anchoring while preserving the ability to exploit highly informative visual evidence when available.

\subsection{Human Adoption Optimization}

In the human-AI competition setting, answer quality alone is insufficient. Prior work has shown that effective human-AI collaboration depends not only on predictive performance but also on the interpretability and trustworthiness of model recommendations \citep{bansal2021does}. Human captains must be able to quickly evaluate and trust the system's recommendations. Consequently, the Bonus agent was explicitly optimized for adoption.

During development, we observed that verbose explanations and hedged responses reduced human willingness to use the model's suggestion. To address this issue, outputs were constrained to concise answer-centered formats with calibrated confidence estimates and short supporting rationales. Confidence scores serve as a lightweight uncertainty signal, while explanations provide sufficient evidence for rapid auditing without overwhelming the user.

\subsection{Development and Error Analysis}

The final Bonus agent emerged through a series of iterative refinements. Early versions often produced overly verbose explanations that negatively affected adoption. Subsequent revisions emphasized concise answer retrieval, confidence calibration, and structured outputs. Additional improvements focused on reducing entity confusion by prioritizing distinctive clues and delaying commitment when multiple closely related candidates remained plausible.

A second class of failures involved multimodal reasoning and quantitative questions. These issues were addressed through evidence routing, cross-modal verification, and the incorporation of domain-specific reasoning policies inherited from the overall system design. The resulting agent achieved a Bonus Effect score of 0.164 and contributed substantially to the highest overall leaderboard score obtained during the competition.

\section{Experiments and Results}
\label{sec:results}

\subsection{Experimental Setup}

Experiments were conducted within the official QANTA 2026 shared challenge hosted as part of the ICML 2026 Workshop on Efficient Multimodal Question Answering (EMM-QA). The competition evaluates hosted multimodal question-answering systems on both Tossup and Bonus tasks. Systems receive streaming textual clues and, for selected questions, accompanying images. Performance is evaluated using task-specific metrics that jointly capture answer quality, confidence calibration, buzzing behavior, and human-AI collaboration effectiveness. The results reported in this paper correspond to the official challenge submissions \texttt{QANTA41Mini\_V1} for Tossup and \texttt{QANTA41\_Bonus\_V1} for Bonus.

\subsection{Tossup Results}

Table~\ref{tab:tossup_results} summarizes the performance of the proposed Tossup agent. The system achieved a Tossup score of 0.238, the highest score on the leaderboard at evaluation time. 

\begin{table}[ht]
\centering
\caption{Final Tossup Results.}
\label{tab:tossup_results}
\begin{tabular}{lc}
\toprule
Metric & Value \\
\midrule
E[Score] & 0.238 \\
Buzz Precision & 72.5\%\\
Buzz Frequency & 94.2\% \\
Buzz Position & 60.558 \\
Win Rate & 71.1\% \\
Total Evaluation Cost (USD) & \$0.14 \\
\bottomrule
\end{tabular}
\end{table}

The agent maintained a Buzz precision of 72.5\%  while buzzing on 94.2\% of questions and achieved a Win Rate of 71.1\%. Despite operating under hosted-agent constraints, the system achieved competitive buzzing behavior at an inference cost of only \$0.14. The reported cost corresponds to the total API expenditure incurred during the official Tossup evaluation set, demonstrating the extreme efficiency of the GPT-4.1-mini backbone for incremental processing.

\subsection{Bonus Results}

Table~\ref{tab:bonus_results} presents the performance of the Bonus agent. The system achieved a Bonus Effect of 0.164 together with 89.1\% Part Accuracy and 72.7\% Question Accuracy.

\begin{table}[ht]
\centering
\caption{Final Bonus Results.}
\label{tab:bonus_results}
\begin{tabular}{lc}
\toprule
Metric & Value \\
\midrule
Bonus Effect & 0.164 \\
Part Accuracy & 89.1\% \\
Question Accuracy & 72.7\% \\
Calibration & 88.2\% \\
Adoption & 33.8\% \\
\bottomrule
\end{tabular}
\end{table}

The agent additionally achieved strong Calibration (88.2\%) and Adoption (33.8\%), indicating that its responses were both reliable and useful in the human-AI collaboration setting. Adoption measures the frequency with which human participants accepted the AI recommendation under the official evaluation protocol.

\subsection{Overall Leaderboard Performance}

Table~\ref{tab:overall_results} summarizes the final leaderboard performance. Combining the Tossup and Bonus agents yielded an Overall Score of 0.402, corresponding to the top-ranked position on the competition leaderboard at evaluation time. The closest competing system obtained an Overall Score of 0.370, resulting in a margin of 0.032.

\begin{table}[ht]
\centering
\caption{Overall Leaderboard Performance.}
\label{tab:overall_results}
\begin{tabular}{lc}
\toprule
Metric & Value \\
\midrule
Overall Rank & 1 \\
Overall Score & 0.402 \\
Tossup E[Score] & 0.238 \\
Bonus Effect & 0.164 \\
Bonus Part Accuracy & 89.1\% \\
Bonus Adoption & 33.8\% \\
\bottomrule
\end{tabular}
\end{table}

These results suggest that task-specific reasoning policies, confidence calibration, and efficient multimodal evidence integration can provide a substantial advantage in resource-constrained multimodal question answering.

\section{Error Analysis and Discussion}
\label{sec:analysis}

Although the proposed system achieved the highest overall leaderboard score, several recurring failure modes were observed during development and evaluation. Analyzing these failures proved critical for improving expected score, confidence calibration, and multimodal reasoning performance. This section discusses the primary sources of error, the design modifications introduced to address them, and the broader lessons learned from developing efficient hosted multimodal question-answering  agents for QANTA 2026.

\subsection{Confidence Calibration}
\label{sec:confidence_calibration}

Confidence calibration emerged as one of the most important factors affecting Tossup performance. Confidence calibration has been widely recognized as an important component of reliable machine learning systems, particularly when predictions are used to drive downstream decisions under uncertainty \citep{guo2017calibration}. In the QANTA setting, incorrect buzzes incur explicit penalties, making overconfident predictions substantially more costly than delayed but correct answers. Consequently, the objective is not simply to maximize answer accuracy, but rather to maximize expected score through calibrated decision-making.

During early development, the system employed more aggressive buzzing thresholds. While these configurations occasionally produced earlier correct answers, they also resulted in frequent incorrect commitments to semantically related entities. Representative examples included confusion between \emph{Frodo} and \emph{Bilbo}, \emph{Nick Carraway} and \emph{Jay Gatsby}, and \emph{Mormonism} and \emph{Catholicism}. In these cases, the model successfully identified the correct semantic neighborhood but lacked sufficient evidence to distinguish among nearby candidates.

Table~\ref{tab:threshold_ablation} summarizes the impact of confidence threshold selection during development using a small set of playground Tossup questions. Although these experiments do not reflect the official leaderboard evaluation, they were useful for understanding the trade-off between buzzing aggressiveness and precision. Lower thresholds increased buzzing frequency but substantially reduced precision and expected score due to incorrect early commitments. In contrast, more conservative thresholds reduced catastrophic errors and improved expected value, motivating the final confidence-gating strategy used in the submitted system.

\begin{table}[ht]
\centering
\caption{Effect of confidence threshold on a qualitative development set of playground Tossup questions.}
\label{tab:threshold_ablation}
\begin{tabular}{lccc}
\toprule
Threshold & Precision & Buzz Position & Expected Score \\
\midrule
0.75 & 25\% & 25 & -0.2 \\
0.85 & 75\% & 39 & 0.0 \\
0.90 & 100\% & 48 & +0.5 \\
\bottomrule
\end{tabular}
\end{table}

A key observation was that confidence estimates generated by GPT-4.1-mini were not uniformly reliable across all ranges. Confidence values 0.80-0.85 frequently corresponded to plausible but unstable hypotheses that were later revised as additional clues became available. By contrast, confidence values above 0.90 were significantly more likely to correspond to the final correct answer. This observation motivated the final confidence threshold used in the submitted system.

The final submitted system achieved a Buzz Precision of 72.5\% on the official Tossup leaderboard, indicating that the trends observed during development largely transferred to the full competition setting. Overall, these experiments demonstrate that confidence calibration is a critical component of incremental question answering systems. Accurate answer generation alone is insufficient; systems must also determine when available evidence is strong enough to justify an irreversible buzzing decision.

\subsection{Related-Entity Confusion}

A recurring challenge in incremental question answering is distinguishing between closely related entities that share substantial contextual overlap. In quizbowl, early clues are intentionally obscure and often describe a broader narrative, historical period, scientific domain, or thematic category before revealing uniquely identifying information. As a result, multiple plausible candidates may remain consistent with the available evidence for a significant portion of the question.

This behavior was particularly evident in literary, historical, and religious questions. For example, clues describing a fictional narrative may initially support several characters from the same work, while clues about a religious movement may remain compatible with multiple denominations until a distinctive identifier appears. In such cases, the model often identified the correct answer family early but lacked sufficient evidence to confidently distinguish among neighboring entities.

We found that generic clues were substantially less informative than distinctive clues. Early commitment based on thematic similarity frequently produced unstable hypotheses, whereas rare names, unique events, specific quotations, and highly distinctive attributes were much stronger predictors of the final answer. This observation motivated the final prompting strategy, which emphasized distinctive evidence and encouraged uncertainty when multiple related candidates remained plausible.

These findings suggest that a significant portion of quizbowl difficulty arises not from knowledge gaps but from fine-grained entity disambiguation under incomplete information. Improving robustness to related-entity confusion remains an important direction for future multimodal question-answering systems.

\subsection{Lessons Learned}

Several practical lessons emerged during system development. First, confidence calibration proved more important than raw answer accuracy in the Tossup task because incorrect buzzes incur explicit penalties. Second, simpler prompting strategies consistently outperformed heavily engineered reasoning protocols for both Tossup and Bonus agents. Third, multimodal evidence was most effective when used selectively for disambiguation rather than as a primary source of information. Finally, domain-specific guardrails, such as the Numeric Firewall used for quantitative questions, improved robustness by preventing overconfidence from limited numerical evidence.

These observations suggest that efficient multimodal question answering benefits from task-specific reasoning policies, calibrated decision-making, and lightweight evidence integration strategies rather than increasingly complex prompting pipelines.

\section{Conclusion}

This paper presented a task-specific multimodal question-answering system for the QANTA 2026 shared challenge. Rather than relying on a single unified strategy, the proposed approach employed separate agents for the Tossup and Bonus tasks, allowing each component to optimize for its respective objective. The Tossup agent combined GPT-4.1-mini with confidence calibration, expected-value-based buzzing, and domain-specific reasoning guardrails, while the Bonus agent utilized GPT-4.1 with leadin-aware reasoning and multimodal evidence routing.

Experimental results demonstrated that these design choices were effective under hosted-agent resource constraints. The final system achieved a Tossup E[Score] of 0.238, a Bonus Effect of 0.164, and an Overall Score of 0.402, securing first place on the competition leaderboard at the time of submission. Error analysis further highlighted the importance of confidence calibration, selective multimodal fusion, and robust entity disambiguation for incremental question answering.

Overall, our findings suggest that efficient multimodal question answering benefits more from calibrated decision-making, task-specific reasoning policies, and targeted evidence integration than from increasingly complex prompting pipelines. Future work may explore improved entity disambiguation, stronger quantitative reasoning capabilities, and adaptive confidence estimation methods for incremental multimodal question answering.

\bibliographystyle{icml2026}
\bibliography{references}

\clearpage
\appendix

\section{System Design and Reproducibility Details}

This appendix consolidates implementation and design details referenced throughout the paper to improve transparency, reproducibility, and understanding of the proposed system. It summarizes the model backbones, agent responsibilities, reasoning policies, evaluation conditions, and additional competition statistics referenced throughout the manuscript.

\subsection{Model Backbones}

The proposed system employs two specialized language-model agents.

\begin{itemize}
    \item \textbf{GPT-4.1-mini} serves as the backbone for the Tossup task due to its favorable latency, efficiency, and ability to process incrementally revealed clues in real time.
    
    \item \textbf{GPT-4.1} serves as the backbone for the Bonus task, providing stronger reasoning capabilities for multi-part questions, evidence integration, and multimodal analysis.
\end{itemize}

Both agents operated within the official hosted QANTA 2026 evaluation environment.

\subsection{Agent Roles}

The system consists of two task-specific agents designed for complementary responsibilities. Glint and Guild denote the official Tossup and Bonus agents used in the competition submissions, respectively.

\paragraph{Glint (Tossup Agent).}
Glint is responsible for incremental clue processing, answer prediction, confidence estimation, and buzzing decisions. As new clues become available, the agent continuously updates candidate hypotheses and issues a buzz only when the confidence threshold is satisfied.

\paragraph{Guild (Bonus Agent).}
Guild is responsible for bonus-question answering, evidence aggregation, multimodal reasoning, and answer refinement. The agent generates concise answer-centered responses intended to support human decision-making during the Bonus phase.

\subsection{Prompting Principles}

Prompt design was guided by three primary principles:

\begin{itemize}
    \item \textbf{Distinctive Clue Prioritization}: Encourage reasoning based on highly discriminative evidence rather than generic associations.
    
    \item \textbf{Concise Answer-Centered Outputs}: Reduce unnecessary generation while improving readability and usability for human participants.
    
    \item \textbf{Leadin-Aware Reasoning}: Utilize question lead-ins to establish the expected semantic domain before evaluating candidate answers.
\end{itemize}

\subsection{Confidence Policy}

The Tossup agent produces both an answer prediction and a verbalized confidence estimate, denoted as $P(\text{correct})$.

Buzzing decisions are governed by a confidence threshold of 0.90. This threshold was adopted to prioritize precision and reduce incorrect commitments during incremental clue revelation. Confidence estimates are treated as decision-support signals rather than formally calibrated probabilities.

\subsection{Numeric Firewall}

To mitigate overconfidence on mathematics and science questions, the system employs a \textbf{Numeric Firewall}.

The Numeric Firewall functions as a prompt-level reasoning guardrail. Confidence is not permitted to increase substantially based solely on isolated numerical values. Numerical evidence must be supported by at least one contextual anchor:

\begin{itemize}
    \item Named entities,
    \item Explicit formulaic context,
    \item Multiple independent and consistent clues.
\end{itemize}

This policy reduces premature commitment to superficially plausible numerical answers and improves confidence calibration in quantitative domains.

\subsection{Multimodal Reasoning Policy}

The system employs a late-fusion multimodal reasoning strategy.

Textual clues are first used to generate candidate hypotheses. Visual evidence is subsequently used as a secondary verification mechanism. Images may contribute OCR text, diagrams, artistic styles, or contextual cues that support, refine, or contradict text-derived hypotheses.

Final predictions are determined through cross-checking between textual and visual evidence streams.

\subsection{Evaluation Environment}

All experiments were conducted within the official hosted-only QANTA 2026 evaluation environment.

The Tossup task utilized incrementally revealed textual clues. The Bonus task incorporated textual evidence and, when available, visual information supplied by the competition platform.

All reported leaderboard results correspond to the official evaluation framework provided by the organizers.

\subsection{Competition Statistics}

Table~\ref{tab:competition_stats} summarizes additional statistics reported during the official QANTA 2026 evaluation.

\begin{table}[h]
\centering
\caption{Additional competition statistics for the proposed system.}
\label{tab:competition_stats}
\begin{tabular}{lcc}
\toprule
\textbf{Agent} & \textbf{Metric} & \textbf{Value} \\
\midrule
Glint (Tossup) & Tossup Wins & 37 \\
Glint (Tossup) & Tossup Accuracy & 78.4\% \\
Guild (Bonus) & Bonus Accuracy & 71.2\% \\
\bottomrule
\end{tabular}
\end{table}

These statistics provide additional insight into the complementary strengths of the two agents. Glint achieved the highest number of Tossup wins among participating systems, while Guild maintained strong Bonus performance through evidence aggregation and multimodal reasoning. Together, the agents contributed to the system achieving the highest overall leaderboard score in the QANTA 2026 Human--AI Hybrid Quiz Bowl Challenge.

\end{document}